\documentclass[letterpaper, 10 pt, conference]{ieeeconf}  

\IEEEoverridecommandlockouts                              

\usepackage[pagebackref=true,breaklinks=true,colorlinks,bookmarks=false]{hyperref}      
\usepackage{graphicx}
\usepackage[utf8]{inputenc}
\usepackage{amssymb}
\usepackage{threeparttable}
\usepackage{amsmath}
\usepackage{booktabs}
\usepackage{multirow}
\usepackage{subcaption}
\usepackage[table]{xcolor} 
\usepackage{hyperref}

\title{\LARGE \bf
KGN-Pro: Keypoint-Based Grasp Prediction through Probabilistic 2D-3D Correspondence Learning
}

\author{Bingran Chen$^{1}$, Baorun Li$^{1}$, Jian Yang$^{5}$, Yong Liu$^{1,4}$, and Guangyao Zhai$^{2,3\dagger}$ 
\thanks{$^{1}$Zhejiang University \quad $^{2}$Technical University of Munich \quad $^{3}$Munich Center for Machine Learning \quad $^{4}$The State Key Laboratory of Industrial Control Technology \quad $^{5}$China Research and Development Academy of Machinery Equipment \quad $^\dagger$Corresponding Author (e-mail: {\tt guangyao.zhai@tum.de})}\
}

\begin{document}

\maketitle

\thispagestyle{empty}
\pagestyle{empty}

\begin{abstract}
High-level robotic manipulation tasks demand flexible 6-DoF grasp estimation to serve as a basic function. Previous approaches either directly generate grasps from point-cloud data, suffering from challenges with small objects and sensor noise, or infer 3D information from RGB images, which introduces expensive annotation requirements and discretization issues. Recent methods mitigate some challenges by retaining a 2D representation to estimate grasp keypoints and applying Perspective-n-Point (PnP) algorithms to compute 6-DoF poses. However, these methods are limited by their non-differentiable nature and reliance solely on 2D supervision, which hinders the full exploitation of rich 3D information.
In this work, we present KGN-Pro, a novel grasping network that preserves the efficiency and fine-grained object grasping of previous KGNs while integrating direct 3D optimization through probabilistic PnP layers. KGN-Pro encodes paired RGB-D images to generate Keypoint Map, and further outputs a 2D confidence map to weight keypoint contributions during re-projection error minimization. By modeling the weighted sum of squared re-projection errors probabilistically, the network effectively transmits 3D supervision to its 2D keypoint predictions, enabling end-to-end learning. Experiments on both simulated and real-world platforms demonstrate that KGN-Pro outperforms existing methods in terms of grasp cover rate and success rate. Project website:   \href{https://waitderek.github.io/kgnpro}{\texttt{https://waitderek.github.io/kgnpro}}.
\end{abstract}
\section{Introduction}
While deep learning has advanced top-down grasp detection, high-level robot manipulation tasks require 6-DoF grasp estimation for greater flexibility, including object rearrangement~\cite{zhai2024sg,liu2022structformer} and vision-language robotic control~\cite{tang2023graspgpt,rashid2023language}.  A typical 6-DoF system first generates candidate grasps, then ranks them based on factors like grasp quality, collision risks, and environmental constraints. Ultimately, a robust grasp generation algorithm must produce accurate and diverse candidates to ensure effective operation in constrained settings.

\begin{figure}[t]
    \centering
    \includegraphics[width=0.9\linewidth]{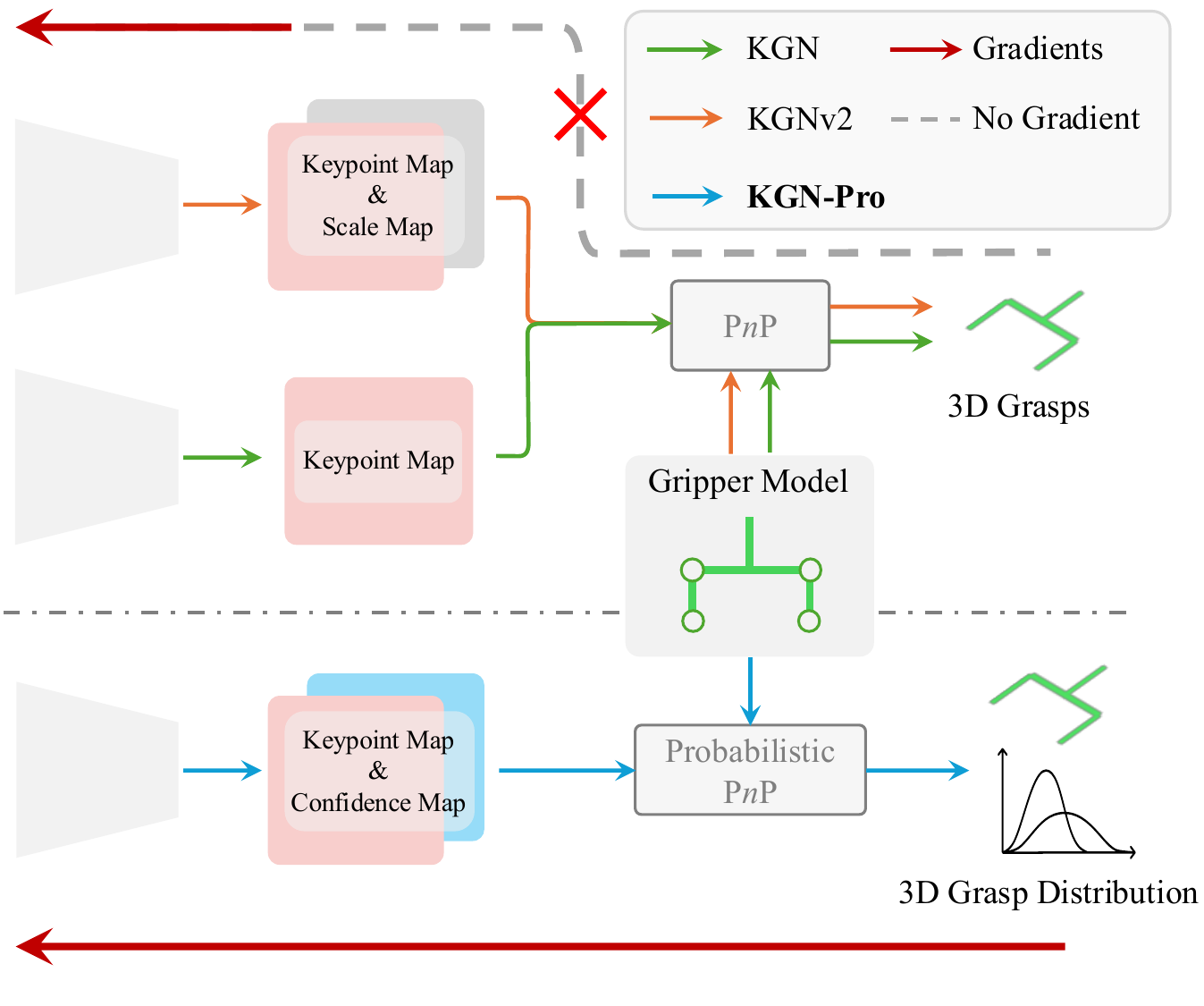}
    \caption{Schematic comparison between previous KGNs and the proposed KGN-Pro. KGNs only supervise Keypoint Map, as the traditional P$n$P blocks the gradients backpropagating to the image encoder. In contrast, KGN-Pro models a grasp distribution by probabilistic P$n$P, guaranteeing end-to-end training in the 3D space.}
    \label{teaser}
    \vspace{-5mm}
\end{figure}

Point-cloud approaches using PointNets~\cite{qi2017pointnet,qi2017pointnet++} generate grasps directly from 3D perception but struggle with small objects and sensor noise. Although a point sampling strategy~\cite{ma2023towards} can balance object scales, it increases computational cost due to the additional segmentation network, and sensitivity to input disturbances remains an issue.
RGB-based input for 6-DoF grasp detection is gaining traction because color images offer richer visual details. CNN-extracted RGB cues help detect small objects that depth data might miss and improve noise robustness. However, methods~\cite{gou2021rgb,zhai2023monograspnet} still relying on 3D grasp representations force networks to infer 3D information from 2D/2.5D inputs. This dependency leads to the need for expensive annotations, such as surface normals or heavy discretization of the $\mathrm{SO}(3)$ space.
In contrast, the KGN series~\cite{chen2023keypoint,chen2023kgnv2} retains a 2D representation by first estimating grasp Keypoint Map—drawing inspiration from planar grasping~\cite{xu2022gknet}—and then using a Perspective-n-Point (P$n$P) algorithm to compute 6-DoF grasp poses based on 2D-3D correspondences. Because P$n$P is non-differentiable, KGN~\cite{chen2023keypoint} cannot directly incorporate 3D supervision and must rely solely on 2D keypoint optimization, as shown in Fig.~\ref{teaser}. This reliance on 2D supervision limits the method's ability to fully exploit rich 3D information, thereby constraining its overall effectiveness.
An intuitive idea is to bypass the P$n$P problem by building a 3D network whose output is directly supervised by 3D grasp points. However, networks suffer from scaled geometric number prediction and tend to behave better when predicting residuals or increments, as evidenced in~\cite{wang2024glace}. This is also a shared drawback of KGNv2~\cite{chen2023kgnv2}, which predicts grasp scales to adjust grasp poses when solving P$n$P.

In this paper, we introduce \textit{KGN-Pro}, a novel grasping network that incorporates direct 3D optimization via probabilistic P$n$P layers to significantly enhance grasp prediction performance without scale prediction problems while preserving the advantages of previous KGNs: fast data processing, pure 2D representations, and fine-grained object grasping.
Specifically, \textit{KGN-Pro} first encodes a pair of RGB-D images to obtain the heatmaps of grasping centers and the offsets to each keypoint. 
In addition, \textit{KGN-Pro} outputs a 2D confidence map to indicate the reliability of each keypoint during the P$n$P re-projection error calculation.
This confidence map ensures effective propagation of 3D supervision signals to the Keypoint Map, thereby enabling end-to-end learning, while also filtering keypoints during inference, generating grasping poses with enhanced accuracy
In contrast to object pose estimation, where a single pose is assigned per object, our approach generates multiple grasp candidates and employs a multimodal matching method to determine the most appropriate optimization direction and target.
Based on this framework, we model the minimization of the weighted sum of squared re-projection errors as a probabilistic problem, considering all possible poses and computing the probability density distribution. 
The \textit{Nearest-Neighbor Matching} is then used to find the most suitable ground truth pose, and the Monte Carlo pose loss is calculated to measure the difference between the predicted and ground truth distributions.
Experiments on both simulation and real-world platforms demonstrate that \textit{KGN-Pro} outperforms competing methods in terms of grasping cover rate and success rate. 
For example, in single-object scenes our method achieves nearly 96\% success under some conditions and maintains robust performance under stringent precision requirements, while in multi-object scenes it consistently exceeds the performance of alternative approaches.

We summarize our contributions as follows:
\begin{itemize}
\item Introduce \textit{KGN-Pro}, a novel grasping network that combines efficient and pure 2D representation with direct 3D optimization using probabilistic PnP layers.
\item Develop a 2D confidence map to effectively weight keypoints during re-projection error minimization, enabling end-to-end training with richer 3D supervision by \textit{Nearest-Neighbor Matching}.
\item Demonstrate through experiments on simulated and real-world platforms that \textit{KGN-Pro} achieves higher grasp cover and success rates compared to state-of-the-art approaches.
\end{itemize}

\section{Related work}
\noindent
\textbf{Direct Supervision}
Direct supervision methods focus on learning grasp poses directly from input data, such as RGB-D images or 3D point clouds~\cite{ten2017grasp}. For instance, Dex-Net~\cite{mahler2016dex} models utilize large-scale datasets and deep neural networks to generate grasp candidates and evaluate their success probabilities. However, these methods often frame the task as a classification problem, which limits their ability to adapt to complex geometric features in cluttered environments.

Similarly, GGCNN~\cite{morrison2018closing} predicts grasp poses at the pixel level, making it suitable for real-time applications with its lightweight architecture. However, its lack of geometric reasoning limits its effectiveness in cluttered scenes~\cite{zhou20176dof, gualtieri2016high}. Methods like GPD~\cite{ten2017grasp}, PointNetGPD~\cite{liang2019pointnetgpd,zhai20222} use 3D point clouds to provide richer geometric information. Still, point clouds lose structural information due to their unordered nature, incur high computational costs in dense scenes, and require error-prone preprocessing, undermining grasp robustness. Unlike these approaches, KGN-Pro improves the efficiency of grasping predictions through compact 2D keypoint representations while avoiding the computational burden of 3D point cloud processing.


\noindent
\textbf{Indirect Supervision}
Using low-cost demonstrations and self-guided reinforcement learning, robots can continuously optimize grasping strategies in diverse environments~\cite{zahavy2020grasping, zeng2018learning, yang2020reinforcement}. 
However, reinforcement learning approaches generally require high computational cost, large-scale data collection, and extended training periods~\cite{pinto2016supersizing, mandikal2021learning}.

To address these challenges, recent approaches leverage intermediate representations (e.g., keypoints, contact points) to decompose grasping tasks and capture local object features, enhancing computational efficiency, robustness, and generalization to unseen objects and complex geometries~\cite{zhai2023monograspnet,chen2023keypoint,sundermeyer2021contact,manuelli2019kpam,  li2015data,kumar2014contact}.
Despite simplifying 3D grasp pose estimation and enhancing computational efficiency, these methods remain dependent on projection accuracy, rendering them susceptible to errors in challenging environments.
KGNs~\cite{chen2023keypoint,chen2023kgnv2} improve the efficiency of keypoint detection and improve the grasping accuracy by detecting the projected gripper keypoints in the image and then using the P$n$P algorithm to recover the 6-DoF grasp pose. However, due to the sensitivity of P$n$P to noise and its non-differentiability, KGN cannot directly incorporate 3D supervisory signals and relies entirely on optimizing 2D keypoints. In contrast, KGN-Pro uses a differentiable probabilistic P$n$P, enabling end-to-end direct 3D supervision, which improves robustness under sensor noise.

%


\begin{figure*}[t]
    \vspace{0.2cm}
    \centering
    \includegraphics[width=\textwidth]{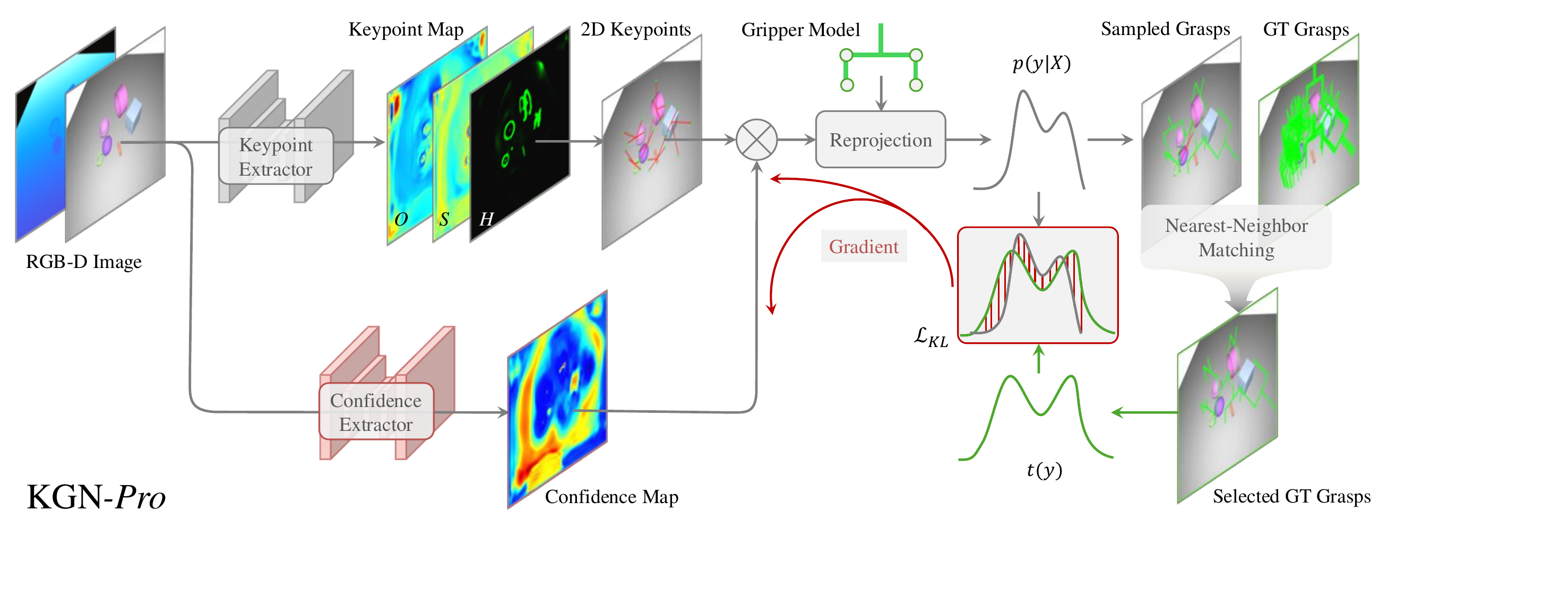}
    \caption{Overview of the proposed KGN-Pro. It takes a pair of RGB-D images as the input and stacks a keypoint extractor to obtain a \textit{Keypoint Map}, where the 2D keypoints can be calculated. Meanwhile, it uses a confidence extractor to obtain the confidence score of each 2D checkpoint, which is considered by 2D-3D correspondences $X$ between 2D keypoints and the corner points on the gripper model. Then, A grasp pose distribution $p(y|X)$ is estimated by the reprojection function on $X$. Finally, it samples grasps from the distribution and performs \textit{Nearest-Neighbor Matching} with ground truth labels to obtain the corresponding pose supervision. The nearest supervision provides a target distribution $t(y)$ to regularize $p(y|X)$.}
    \label{pipeline}
\end{figure*}

\section{Preliminaries}
\label{pre}

P$n$P is a widely used algorithm for pose estimation, where the goal is to determine the optimal pose $y=(R,t)$ by minimizing the reprojection error between the 3D points $p_i^{3D}$ and the observed 2D points $p_i^{2D}$.
To facilitate end-to-end learning, the pose output from the P$n$P algorithm could be modeled as a probability distribution over potential poses, making the entire process differentiable. 
For a given pose $y$ and a specific set of data $x_i=\left\{ p_i^{2D},p_i^{3D}\right\}$, the corresponding probability density function is:

\begin{equation}
    p(x_i | y) = \frac{1}{\sqrt{2\pi\sigma_i^2}} \exp\left( -\frac{\| E_i(y) \|^2}{2\sigma_i^2} \right),
\end{equation}
where $E_i(y)$ represents the re-projection error computed for $x_i$ and $\sigma_i$ is the standard deviation of the error distribution~\cite{chen2022epro}.

The likelihood function for the entire set of observations, $p(X|y)$, where $X$ represents the collection of all predicted 2D and 3D points, can be obtained by multiplying the individual probabilities. 
This likelihood is proportional to the exponentiated negative sum of squared weighted re-projection errors across all data points.
%
Considering $p(y)$ as a non-information prior, we can apply Bayes' Theorem to derive the posterior $p(y|X)$.

To train our model and align $p(y|X)$ with the ground truth pose, we introduce a target pose distribution $t(y)$. Minimizing the KL divergence $D_\text{KL}(t(y)||p(y|X))$ drives the posterior to match the ground truth while naturally accounting for uncertainties. Formally, we obtain the KL divergence loss as:

\begin{equation}
    L_{\text{KL}} = \underbrace{\log \int p(X|y) \, dy}_{L\text{pred}} - \int t(y) \log p(X|y) \, dy + c,
\label{KL_loss}
\end{equation}
where $c$ is a constant. By empirically setting $t(y)$ as a Dirac-like target distribution centered at the ground truth pose $y_{gt}$, the expression can be further simplified.
The first term of Eq.\ref{KL_loss} is approximated using Adaptive Multiple Importance sampling~\cite{cornuet2012adaptive} and each sample $y_j$ is weighted by $v_j$.
For better training efficiency, the proposal distribution $q(y)$ is optimized to enhance the accuracy of the integral approximation.

\begin{equation}
L_{\mathrm{pred}} \approx \log \frac{1}{K} \sum_{j=1}^K \underbrace{\frac{\exp -\frac{1}{2} \sum_{i=1}^N\left\|E_i\left(y_j\right)\right\|^2}{q\left(y_j\right)}}_{v_j \text { (importance weight) }}.
\label{monte_carlo}
\end{equation}

To improve the model's robustness, uncertainty is introduced by using the re-projection error at the target pose, combined with the expected gradient of the re-projection error under the same importance weights through back-propagation. 
The expected gradient of the reprojection error over the predicted pose distribution is approximated by backpropagating each weighted sample in the Monte Carlo pose loss:

\begin{equation}
\frac{\partial L_{\mathrm{pred}}}{\partial(\cdot)}=  \underset{y \sim p(y \mid X)}{\mathbb{E}} \frac{\partial}{\partial(\cdot)} \frac{1}{2} \sum_{i=1}^N\left\|E_i(y)\right\|^2 .
\label{backward2}
\end{equation}

\section{Methodology}

In this section, we first illustrate the fundamental representation of the grasp keypoints, based on which we present the bottlenecks in the previous and hypothesized methods. Then, we introduce our routine and proposed modules to solve the bottleneck. An overview of our method is shown in Fig.~\ref{pipeline}.

\subsection{Grasp Keypoint Estimation}

Each predefined grasp has four keypoints corresponding to the 2D coordinates projected from the 3D corner points of the gripper model. 
Aligned with previous works~\cite{chen2023keypoint, chen2023kgnv2}, we employ a keypoint extractor $f_\theta$ to encode an RGB-D image pair as a \textit{Keypoint Map}, which comprises: (1) the center heatmap $H$ representing the per-pixel probability of being a grasp center; (2) the center sub-pixel offset map $S$; and (3) the center-to-keypoint offset map $O$, which records the displacement between the center and the four keypoints.

During inference, potential centers of grasp are initially selected by evaluating the center heatmap ${H}$, and low-probability points are filtered out using a fixed threshold.
The sub-pixel offset ${S}$ is then combined with the center point coordinates $(u,v)$ to refine the grasp centers. 
These coordinates are further adjusted using the offset ${O}$ to compute the locations of all keypoints associated with the predicted grasp:
\begin{equation}
p_{i,k}^{2D} = \left( \left(u_i,v_i\right) + S_i{\left(\Delta u,\Delta v \right)}\right) + O_{i}^{k} ,k=1,...,4
\label{projection process}
\end{equation}
where $i$ is the $i$-th largest value on the heatmap $H$.

\subsection{Bottlenecks of Supervisions}
The \textit{Keypoint Map} can be either indirectly or directly supervised by the ground truth 3D points. However, both ways contain information bottlenecks that prevent them from being more accurate.

\noindent
\textbf{Non-Differentiable Bottleneck} In the previous work~\cite{chen2023keypoint, chen2023kgnv2}, the generated 2D keypoints are matched with the 3D corner points on the gripper model to obtain grasp poses by solving a P$n$P problem. However, with only four keypoints available, the P$n$P solution is highly sensitive to noise and prone to inaccuracies, which significantly limits the final estimation performance, as evidenced in~\cite{lepetit2009ep}.
Additionally, since the original P$n$P is non-differentiable, KGN series largely consider projected 2D keypoints, and cannot directly incorporate 3D supervision. Even though KGNv2~\cite{chen2023kgnv2} additionally estimates a pose scale to compensate for this problem, the supervision is not fine-grained, which restricts the network's ability to learn fully in the 3D space and limits its accuracy. 
Existing research on pose estimation highlights the importance of rich 3D supervision signals in improving the accuracy of 6-DoF pose estimation~\cite{wang2019normalized,chen2024secondpose}.

\noindent
\textbf{Direct-Estimation Bottleneck}
As P$n$P prevents 3D gradients from flowing directly back through the network, a straightforward idea is to use several MLP layers to regress the 3D poses directly, bypassing the need for 2D-3D correspondences. However, as demonstrated in Section~\ref{Evaluation}, directly regressing 3D signals to generate grasp poses produces suboptimal results. This finding is consistent with prior observations (e.g.,~\cite{wang2024glace}) that networks generally learn more effectively when estimating residuals or increments rather than relying on raw or scaled numerical measurements.

\subsection{2D-3D Correspondence Learning}
To avoid these problems, we propose to estimate a keypoint confidence map, which helps model the minimization of the weighted re-projection errors as a probabilistic problem for 2D-3D correspondence learning. This way of modeling considers all possible poses and computes the probability density distribution.

\noindent
\textbf{Probabilistic P$n$P with Confidence Map}
Inspired by~\cite{chen2022epro, chen2020end}, we use probabilistic P$n$P as the core of {KGN-Pro} to keep the pure 2D keypoint representation and enable the direct backpropagation of 3D supervision signals to the \textit{Keypoint Map}. 
This method effectively integrates ground truth distribution as 3D supervision signals into the process, enhancing the network's ability to better understand and predict 3D poses.

To backpropagate 3D supervision signals, we additionally introduce a confidence extractor to obtain a 2D confidence map consisting of the weight of each 2D point $w_i^{2D}$, which reflects the reliability of 2D-3D correspondences. $w_i^{2D}$ enables more accurate pose estimation considering the variation in confidence in each feature. 
Thereby, with the corner points on the gripper model, we can derive a predicted pose distribution $p(y|X)$.
As mentioned in Sec.~\ref{pre}, we can convert $p(y|X)$ to $p(X|y)$ by Bayes’ Theorem, and we compute the re-projection error $E_i(y)$ and combine it with $w_i^{2D}$ to incorporate the confidence level of each feature point into the re-projection error calculation. We obtain the following expression:

\begin{equation}
    p(X|y) \propto \exp -\frac{1}{2} \sum_{i=1}^N\ \| w_i^{2D} E_i(y)\|^2.
\label{likelihood}
\end{equation}

\noindent
\textbf{Nearest-Neighbor Supervision}
Additionally, since grasp pose estimation differs from object pose estimation, which only has one candidate to solve, we use \textit{Nearest-Neighbor Matching} to ensure that each predicted pose is primarily associated with at most one physically feasible ground truth pose, yielding a target grasp distribution $t(y)$. This initiates the pose learning of a group of grasps.
By substituting Eq.\eqref{monte_carlo} and Eq.\eqref{likelihood} into Eq.\eqref{KL_loss} and incorporating $t(y)$, we derive the new multi-objective loss function:

\begin{equation}
L_{\text{KL}} =\sum_{j=1}^M \left[ L_{\text{pred}}^j +   \left( \frac{1}{2} \sum_{i=1}^N  \big\| w_i^{2D} E_i(\hat{y}_j) \big\|^2 \right)
\right],
\end{equation}
where $M$ is the number of predicted grasp poses. Here, we omit the constant terms related to the target distribution for clearance.
The first loss term $L_{\text{pred}}^j$ represents the normalized log-likelihood of the $j$-th predicted pose distribution. 
The second term accounts for the weighted re-projection error of the $j$-th predicted pose on its corresponding ground truth $\hat{y}_j$, which is the nearest neighbor matched to the predicted pose.
Through this joint optimization, the network is encouraged to learn 2D-3D correspondences that cover multiple plausible poses simultaneously.
Incorporating 2D weights into the probability density function allows the model to prioritize more reliable feature correspondences, which enhances the accuracy of pose predictions.


\subsection{Train Loss}

The proposed network is fully end-to-end trainable, enabling efficient optimization across its components.
As shown in Fig.~\ref{pipeline}, the total loss function comprises a weighted sum of multiple components, each corresponding to different aspects of the network's output.
The \textit{Keypoint Map} has three sub-components, each with its loss: the center heatmap ${H}$ uses a binary focal loss ($L_H$), the center sub-pixel offset map ${S}$ uses the $L_1$ loss ($L_S$), and the center-to-keypoint offset map ${O}$ also uses the $L_1$ loss ($L_O$).
Additionally, the 2D-3D correspondences are optimized through the KL divergence loss. The total loss is defined as:
\begin{equation}
L=\lambda_H L_H+\lambda_S L_S+\lambda_O L_O + \lambda_{\mathrm{KL}} L_{\mathrm{KL}},
\end{equation}
where $\lambda_H, \lambda_O, \lambda_{\mathrm{KL}}$ are the weight for each loss.

%


\section{Experiments}


\begin{figure*}
\vspace{0.2cm}
\centering
    \begin{subfigure}[b]{0.16\linewidth}
        \centering
        \includegraphics[width=\linewidth]{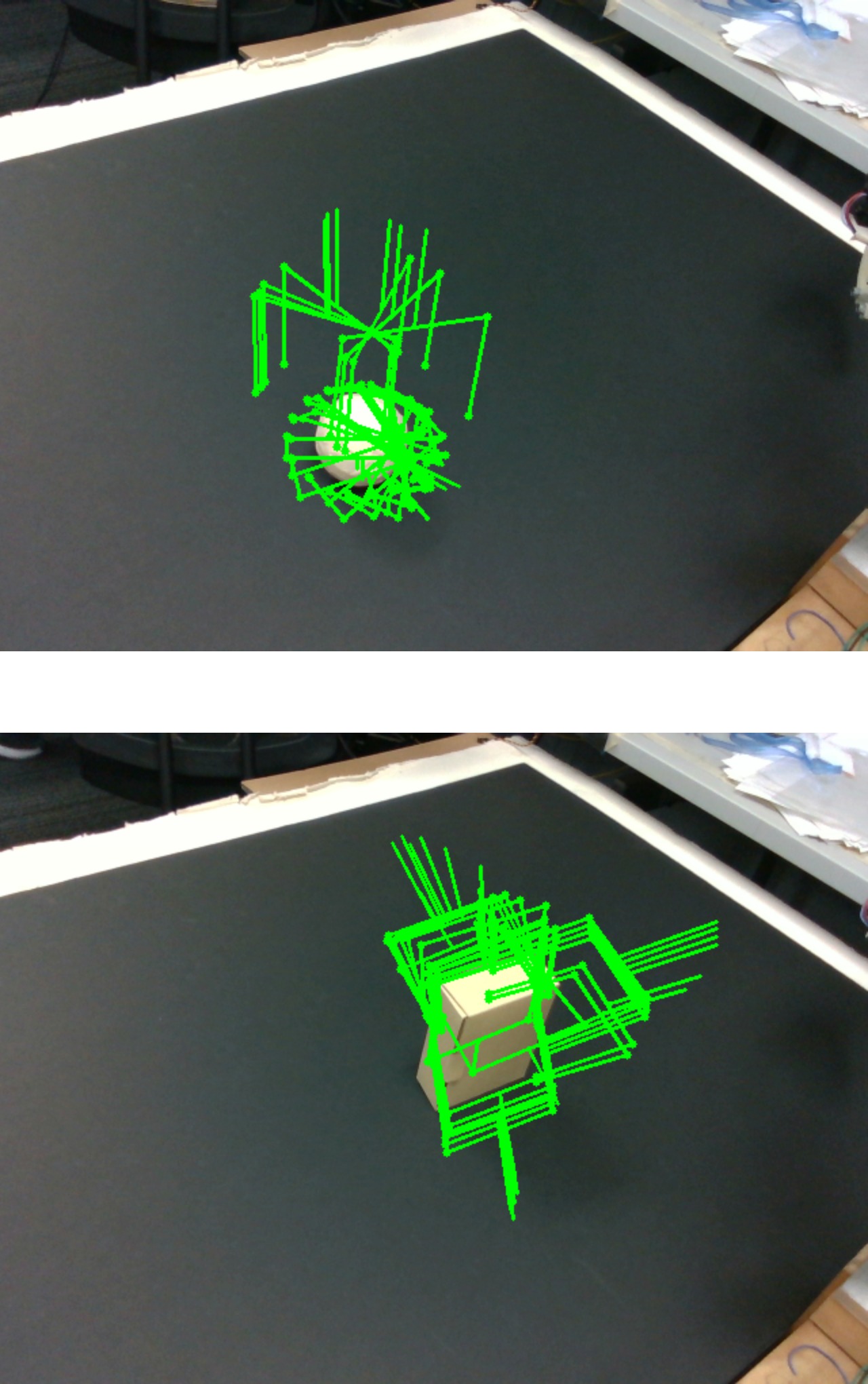}
        \caption{CenterGrasp-single}
        \label{centergrasp_single}
    \end{subfigure}
    \begin{subfigure}[b]{0.16\linewidth}
        \centering
        \includegraphics[width=\linewidth]{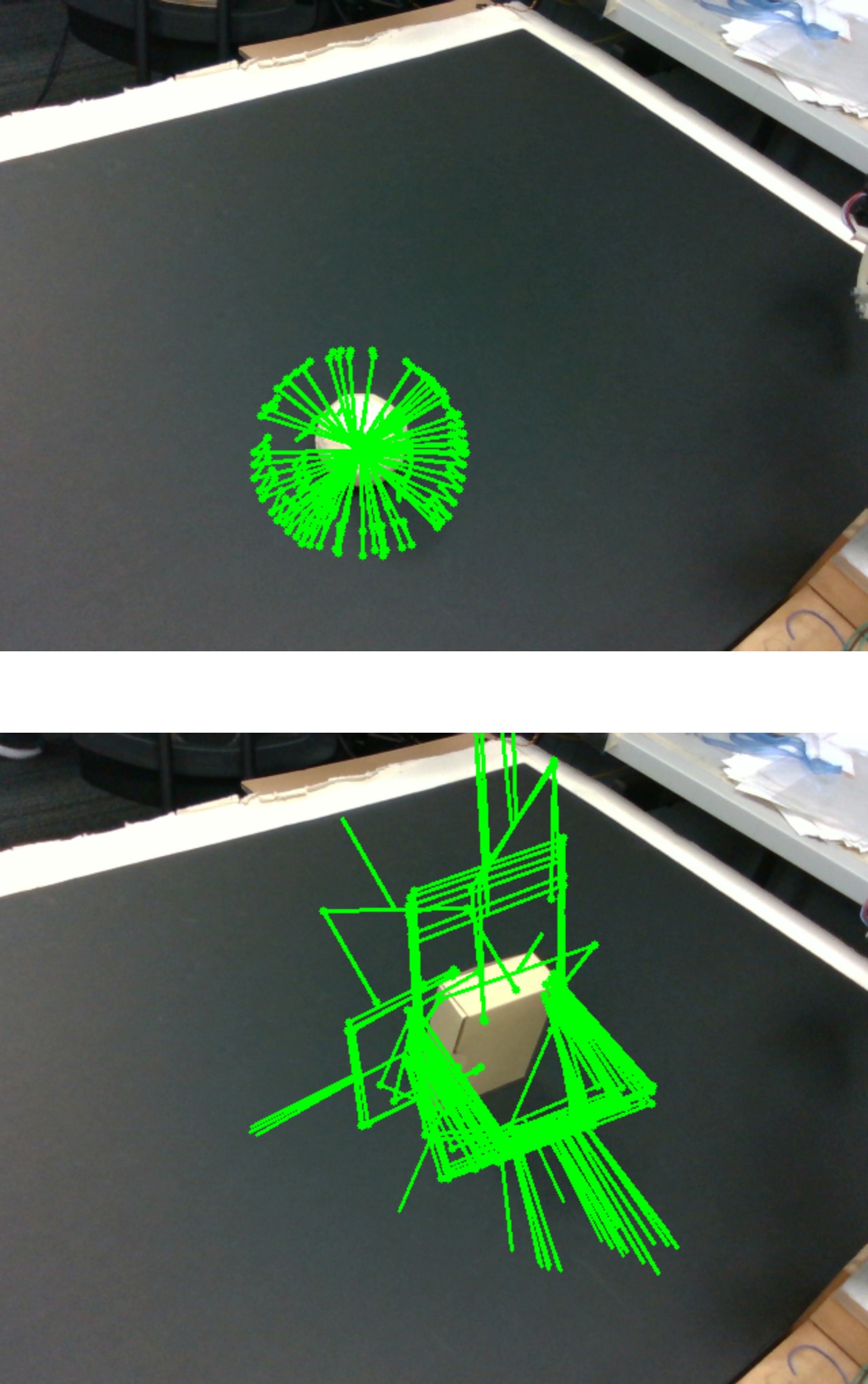}
        \caption{KGN-single}
        \label{KGN_single}
    \end{subfigure}
    \begin{subfigure}[b]{0.16\linewidth}
        \centering
        \includegraphics[width=\linewidth]{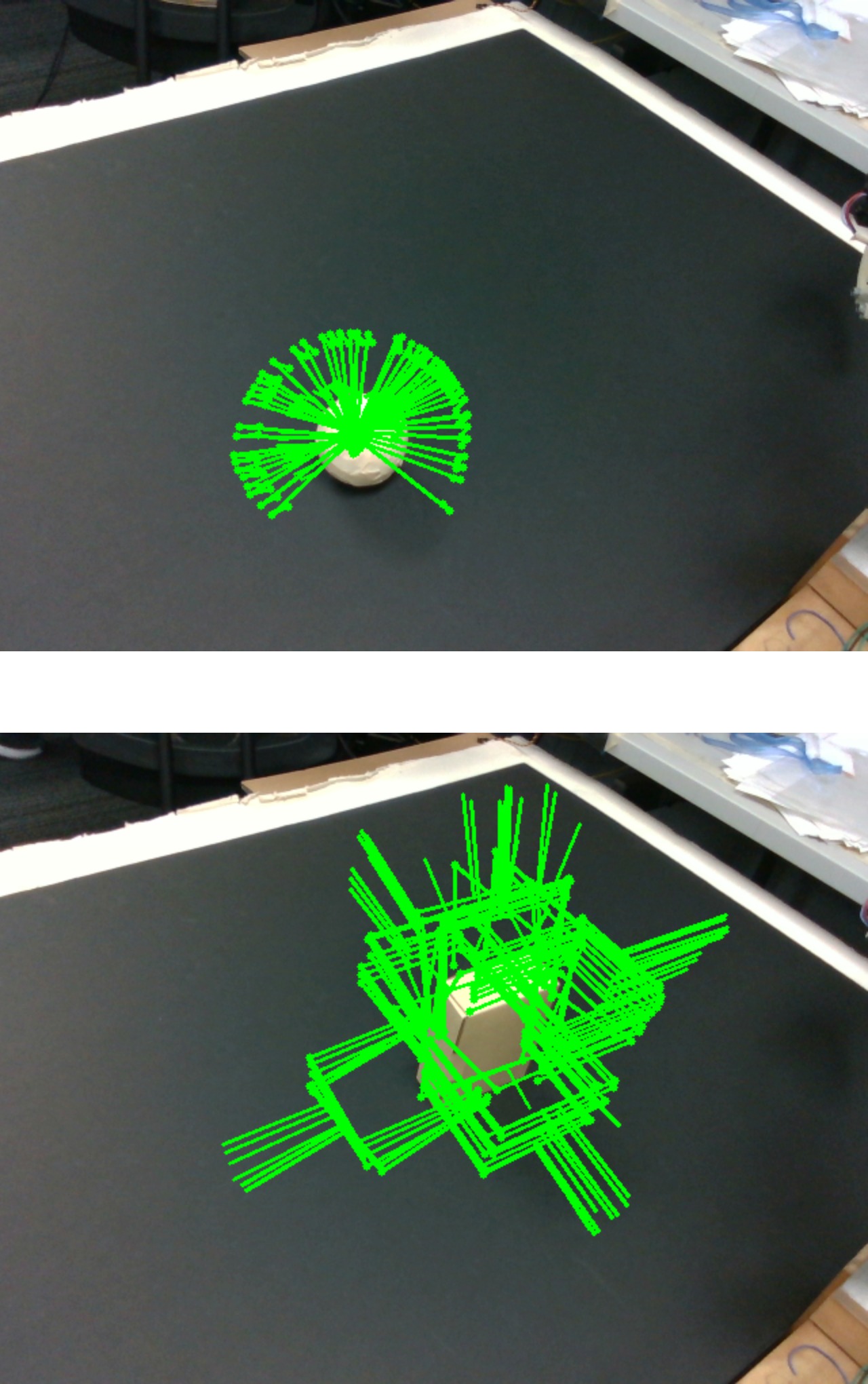}
        \caption{OUR-single}
        \label{OUR_single}
    \end{subfigure}
     \begin{subfigure}[b]{0.16\linewidth}
        \centering
        \includegraphics[width=\linewidth]{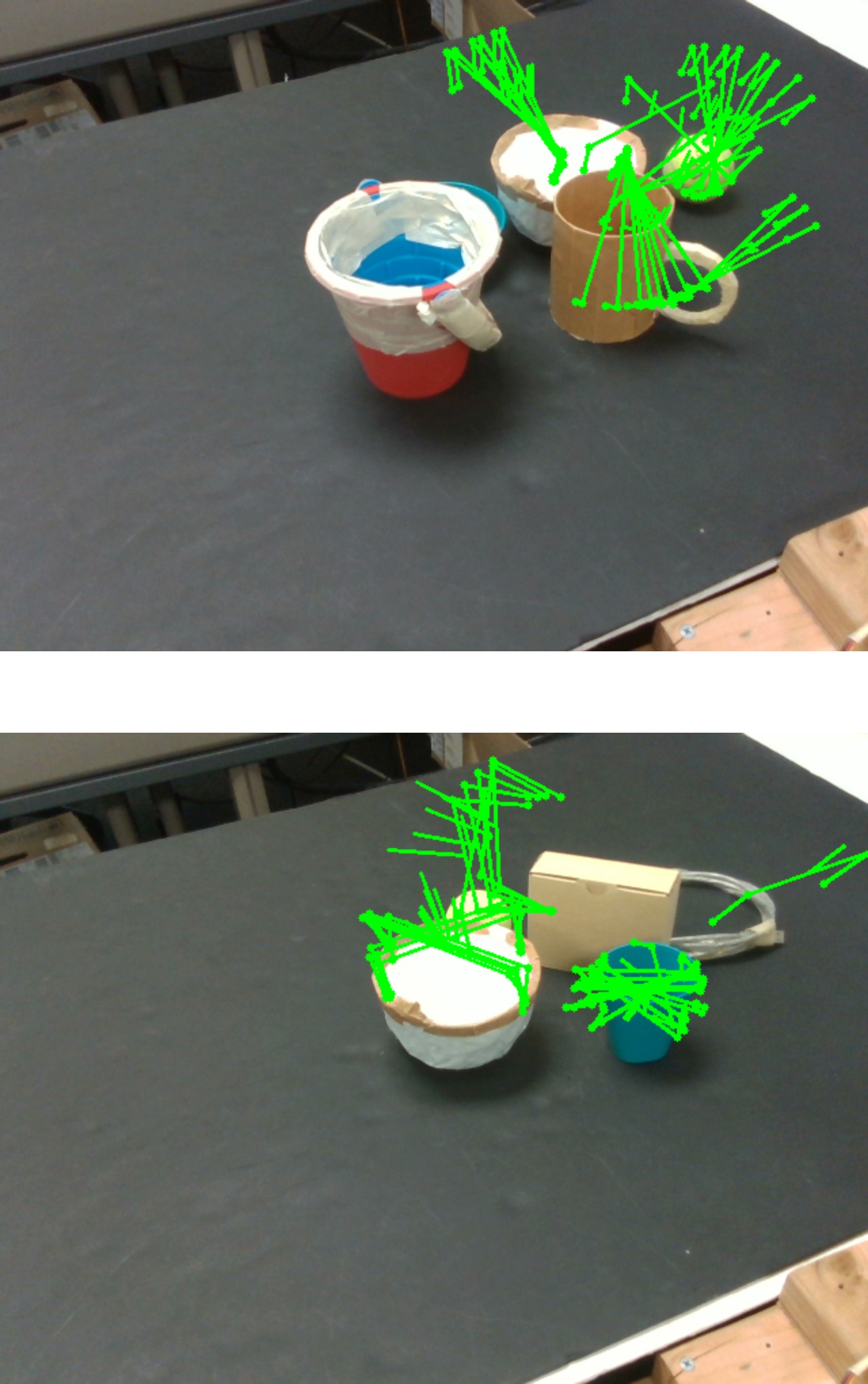}
        \caption{CenterGrasp-multi}
        \label{centergrasp_multi}
    \end{subfigure}
    \begin{subfigure}[b]{0.16\linewidth}
        \centering
        \includegraphics[width=\linewidth]{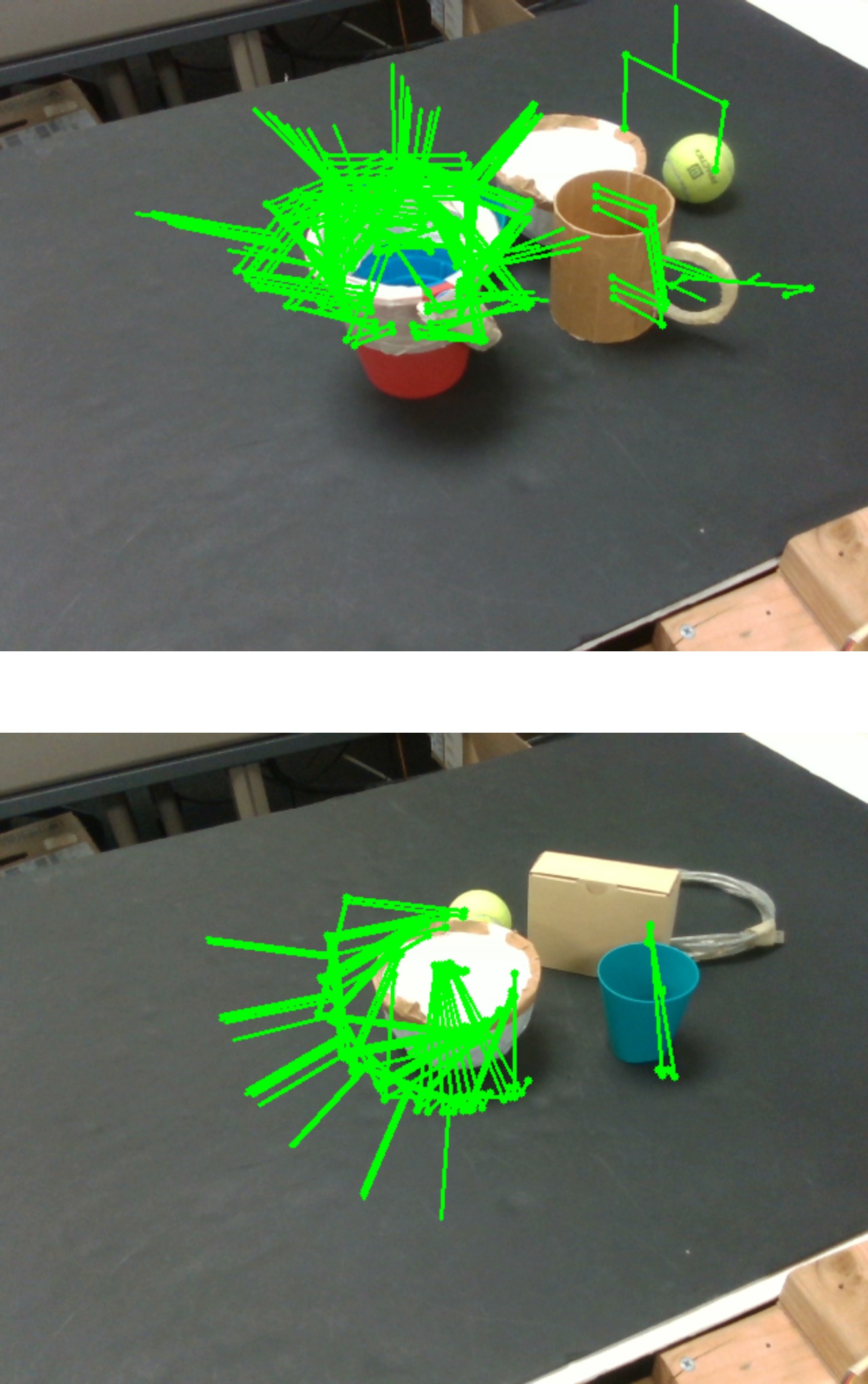}
        \caption{KGN-multi}
        \label{KGN_multi}
    \end{subfigure}
    \begin{subfigure}[b]{0.16\linewidth}
        \centering
        \includegraphics[width=\linewidth]{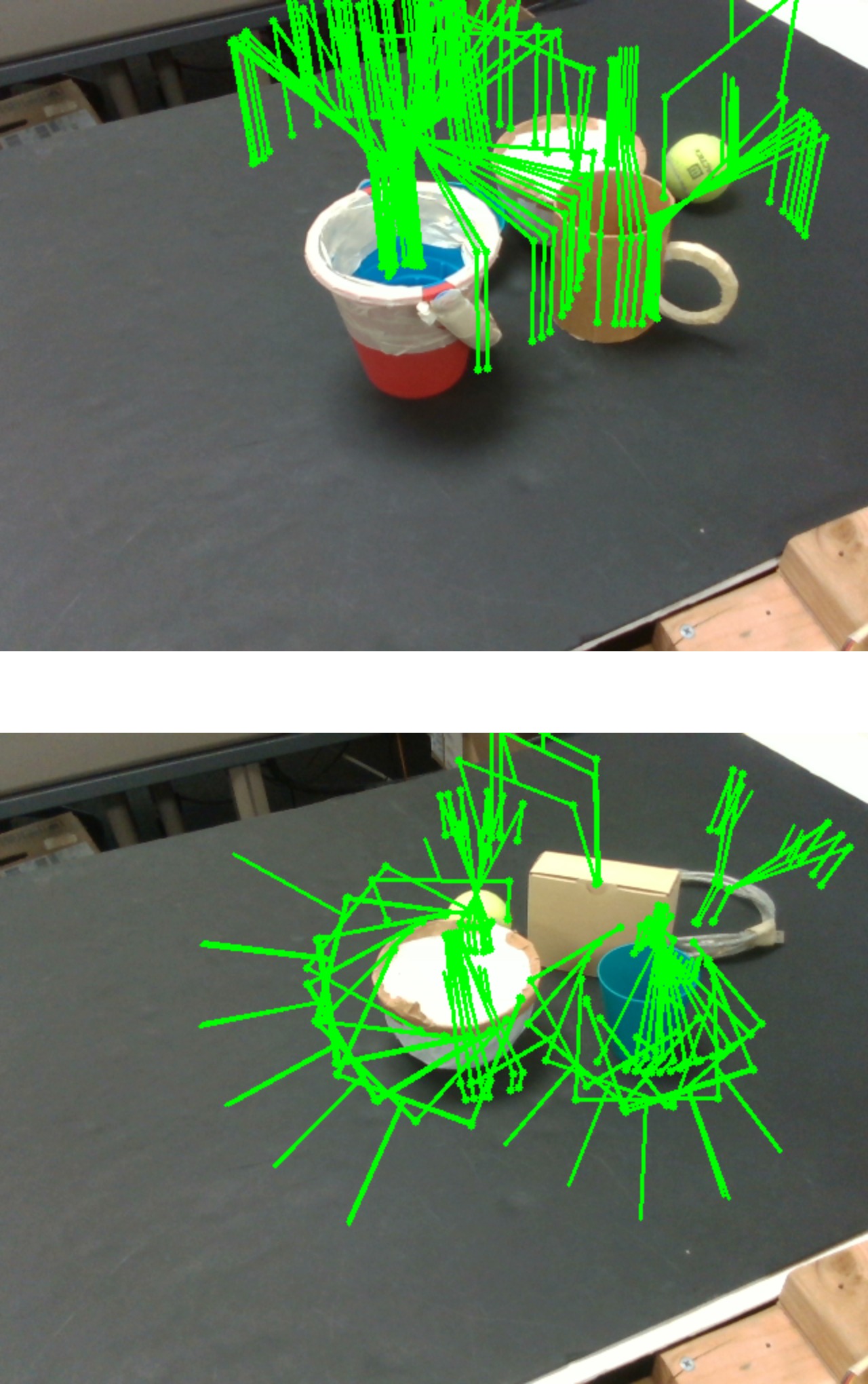}
        \caption{OUR-multi}
        \label{OUR_multi}
    \end{subfigure}
        \caption{Visualization of grasp pose generated by our method, KGN and CenterGrasp. Our method has a clear advantage regarding generated angle diversity and accuracy.}
        \label{totalresult}
\end{figure*}

\begin{figure*}
\centering
    \begin{subfigure}[b]{0.48\linewidth}
        \centering
        \includegraphics[width=\linewidth]{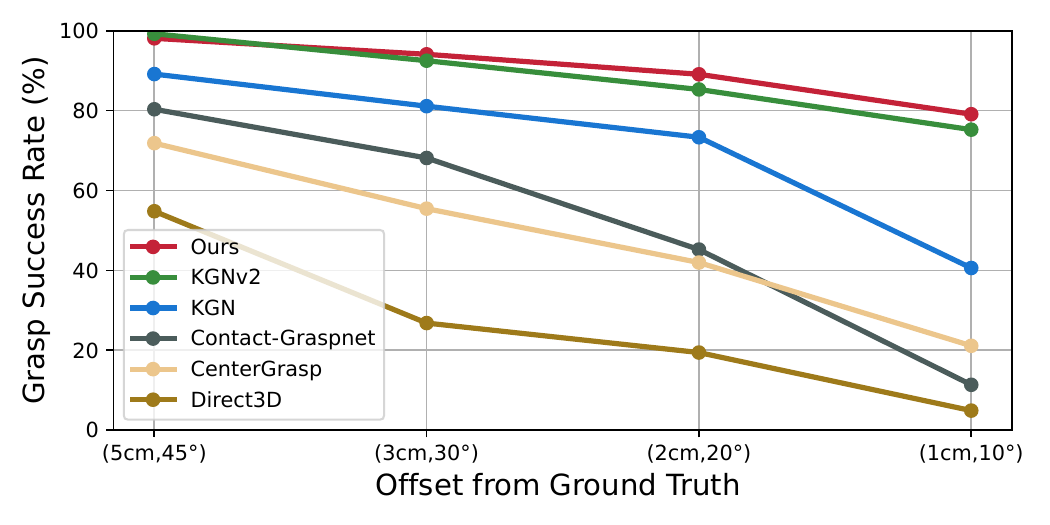}
        \caption{Single-Object Evaluation}
        \label{compare_left}
    \end{subfigure}
    \begin{subfigure}[b]{0.48\linewidth}
        \centering
        \includegraphics[width=\linewidth]{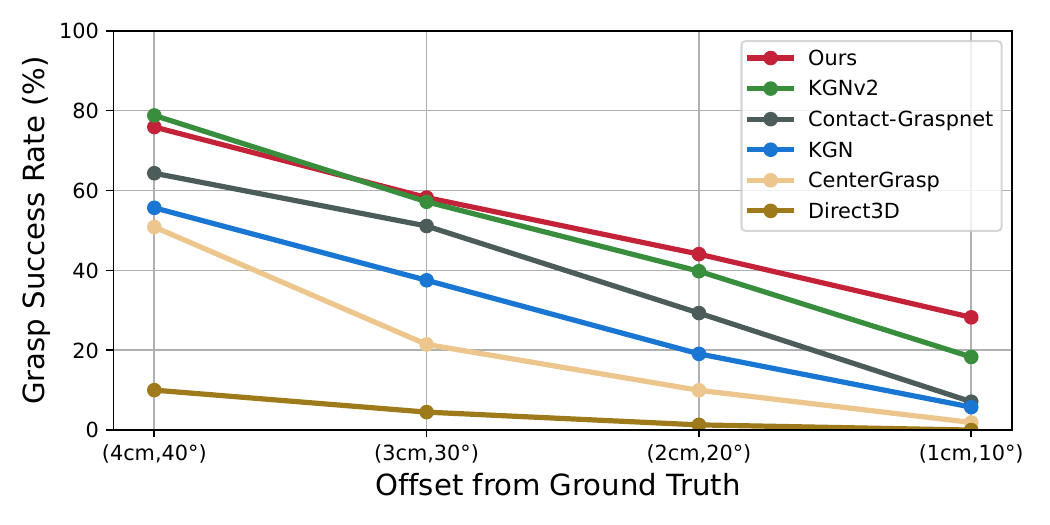}
        \caption{Multi-Object Evaluation}
        \label{compare_right}
    \end{subfigure}
    \caption{All methods are tested on 50 newly generated scenes to verify the generalization ability of the models on unseen data, covering different objects and environmental conditions.}
    \label{single_multi}
\end{figure*}

\subsection{Implementation Details}

\noindent
\textbf{Experiment settings} Our network architecture utilizes the DLA-34~\cite{yu2018deep} backbone as its basic framework, enhanced by deformable convolutional operations~\cite{dai2017deformable}.
Generating multi-scale feature maps at 1/4 of the input dimensions ensures computational efficiency while maintaining rich spatial details.

A well-designed synthetic dataset is essential for the adaptability and robustness of grasping models, as it enhances model generalization across diverse objects and reduces the need for manual annotation~\cite{mousavian20196, yan2018learning, depierre2018jacquard,lin2022primitive, xiang2017posecnn, wen2022catgrasp, viturino20216d}. 
Therefore, we utilized a synthetic dataset, as in KGN~\cite{chen2023keypoint}, to train and evaluate the performance of our improved grasp generation network. 
This dataset spans six representative object categories—Cylinder, Ring, Stick, Sphere, Semi-sphere, and Cuboid—and encompasses both single-target and multi-object scenes, thereby better simulating real-world conditions~\cite{lin2022primitive}.
%
%
The dataset is partitioned into 80\% for training and 20\% for testing, with model training conducted over 300 epochs. The learning rate starts at 1e-4 and is reduced by a factor of 10 at epochs 200 and 250 to facilitate fine-tuning. The weights for each branch of the model are empirically set as follows: $\lambda_Y = 1, \lambda_O = 1, \lambda_J = 1, \lambda_{\mathrm{KL}} = 0.1 $.
This strategy enhances the model's ability to handle extrapolated data while ensuring keypoint prediction remains a primary focus.
As a result, the final grasping poses exhibit improved robustness and generalization in practical applications.

\noindent
\textbf{Baselines.}  We have reproduced two methods representing different approaches on the dataset for the comparison: CenterGrasp~\cite{chisari2024centergrasp} uses point cloud-driven implicit reconstruction to model object geometry and predicts 6-DoF grasp poses. Contact-GraspNet~\cite{sundermeyer2021contact} generates 6-DoF grasps directly from depth-derived point clouds by predicting contact-aligned poses. Additionally, KGNs generate grasp poses by predicting keypoints, and Direct3D supervises the regression of grasp poses directly through 3D signals while using the same backbone as our method. Contact-GraspNet is trained on the Acronym~\cite{eppner2021acronym} dataset, whereas all other methods are trained on the same single-object dataset.

\subsection{Evaluation} \label{Evaluation}
Fig.~\ref{totalresult} illustrates the performance of our method compared to other methods. 
In single-object scenes, our method significantly improves grasp accuracy by generating a broader and more precise range of grasp poses, efficiently covering objects from various angles, and enhancing overall reliability.
This suggests better precision in isolating key features for secure handling. 
While KGN can generate grasp poses across different angles, our method is more reliable in handling complex angles, particularly in challenging scenes.
Especially, as shown in Fig.~\ref{KGN_single} and Fig.~\ref{OUR_single}, our method achieves a wide range of grasping angles, leveraging depth information to effectively avoid collisions.

In the multi-object scenes (right side), the difference between the two methods becomes more evident. 
KGN lacks adequate depth information, which leads to limited flexibility in grasping angles in complex multi-object environments. 
In contrast, our method has a more accurate understanding of space, enabling it to generate a broader range of collision-free grasp postures.


\begin{table}[ht]
\vspace{0.2cm}
\scriptsize
\centering
\renewcommand{\arraystretch}{1.2}
\caption{Object Success Rate(\%)}
\scalebox{1.1}{
\begin{tabular}{llccc}
\toprule
\multirow{2}{*}{{Angle}} & \multirow{2}{*}{{Method}} & {Single-Object} & {Multi-Object} \\
\cmidrule{3-3} \cmidrule{4-4}
\ &  & 1.0 / 1.5 / 2.0 cm& 1.0 / 1.5 / 2.0 cm\\ 
\midrule
\multirow{3}{*}{{10°}} 
 & CenterGrasp~\cite{chisari2024centergrasp} 
 & 20.5 / 44.5 / 56.8  & 13.9 / 33.3 / 50.4\\
 & KGN~\cite{chen2023keypoint}  
 & 60.0 / 71.7 / 78.7  & 69.3 / 73.6 / 83.2\\
 & KGNv2~\cite{chen2023kgnv2} 
 & 91.4 / 97.1 / 97.7 & 84.9 / 87.9 / 89.9\\
 & \textbf{KGN-Pro (Ours)}    
 & \textbf{93.4} / \textbf{96.5} / \textbf{98.5} & \textbf{90.9} / \textbf{93.9} / \textbf{96.9}\\
\midrule
\multirow{3}{*}{{20°}} 
 & CenterGrasp~\cite{chisari2024centergrasp} 
 & 19.5 / 54.5 / 66.8  & 33.5 / 54.5 / 71.8 \\
 & KGN~\cite{chen2023keypoint} 
 & 82.2 / 84.0 / 91.5  & 86.0 / 86.7 / 94.3 \\
 & KGNv2~\cite{chen2023kgnv2} 
 & \textbf{99.4} / \textbf{99.5} / \textbf{100.0} & 90.9 / 93.9 / 96.1\\
 & \textbf{KGN-Pro (Ours)}        
 & 99.1 / \textbf{99.5} / \textbf{100.0} & \textbf{95.2} / \textbf{97.5} / \textbf{98.1} \\
\bottomrule

\end{tabular}
}
\label{table_1}
\end{table}

Fig.~\ref{single_multi} shows that in the single-object scenes, our method achieves a grasp success rate(GSR) of approximately 96\% under the least restrictive conditions (5cm, 45°offset), which is very close to KGNv2.
As the accuracy requirement increases, our method achieves 80\% GSR under the most challenging conditions (1cm, 10° offset), significantly outperforming KGN.
In multi-object scenes, our method's success rate starts at about 80\% and gradually decreases to around 50\% as the accuracy requirement increases. 
It consistently outperforms KGN under all conditions and is slightly better than KGNv2.
This is because the scale-separated detection method proposed by KGNv2 does not directly and explicitly provide 3D supervision signals to help the network understand spatial information.
The performance of CenterGrasp and Contact-Graspnet methods is generally weaker than ours, especially under high precision conditions, where their success rates remain considerably lower.
During our experiments, we observed that the CenterGrasp method produced low-quality point cloud reconstructions in dense object scenes, leading to reduced grasp success rates. 
This is because point cloud-based methods perform poorly in extracting features from small objects, which hinders their grasping accuracy, particularly under high precision conditions.

During the physical experiments, we observed that a large grasping error distance can cause significant deviation of the grasping center from the target object's ideal position.
This may result in improper contact between the manipulator and the object or in the grasping point shifting away from the center of gravity. This leads to instability and ultimately causes grasp failure.
Thus, controlling the error distance within a small range is critical for successful grasping.
As shown in Table~\ref{table_1}, both our method and KGNs achieve a high Object success rate when the error tolerance is set to 2.0 cm. 
However, as the tolerance becomes more stringent, our proposed method demonstrates a clear advantage over both KGN and CenterGrasp, consistently outperforming them in both single-object and multi-object scenes, especially as the angle offset increases. 
These results highlight the effectiveness of our optimization process, providing enhanced reliability and efficiency in high-precision grasping tasks.

\begin{figure}[t]
\vspace{0.2cm}
    \centering
    \includegraphics[width=0.92\linewidth]{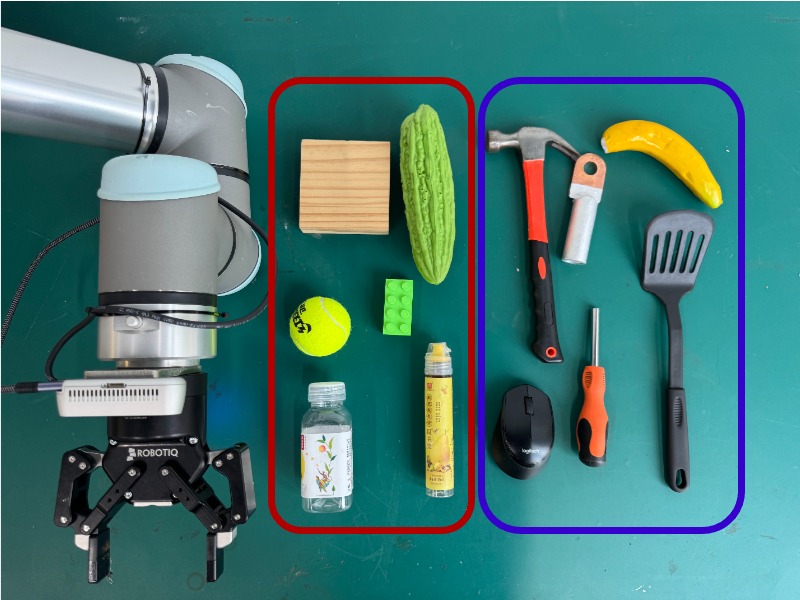}
    \caption{Settings of our physical experiments. The objects within the red box have shapes similar to those found in the dataset, while the objects in the blue box are more geometrically complex.}
    \label{physical}
\end{figure}

\subsection{Physical Experiment}

In the physical experiment, we selected common household items, as shown in Fig.~\ref{physical}, to create both single-object and multi-object scenes for testing.
The ranking of candidate grasping poses was based on two criteria: 2D confidence during detection and the re-projection error after pose recovery. 
The robotic arm was programmed to grasp each object and lift it to a height of 0.1 meters. 
A grasp was considered successful if the object remained securely held during the lifting and placement process.

\begin{table}[ht]
\scriptsize
\centering
\renewcommand{\arraystretch}{1.2}
\caption{The results of physical experiment}
\scalebox{1.1}{
\begin{tabular}{lcccc}
\toprule
\multirow{2}{*}{{Class}} & {Single Scene} & \multirow{2}{*}{{Scene}} & \multicolumn{2}{c}{{Multi Scene}} \\
\cmidrule{2-2} \cmidrule{4-5}
\  & GSR(\%) & \  & GSR(\%) & SCR(\%) \\ 
\midrule
Cylinder & 96\%(24/25) & No.1 & 87.72\%(50/57) & 100.0 (10/10) \\
Sphere & 100\%(25/25) & No.2 & 81.96\%(50/61) & 100.0 (10/10) \\
Stick & 92\%(23/25) & No.3 & 73.84\%(48/65) & 90.0 (9/10) \\
Cuboid & 92\%(23/25) & No.4 & 63.89\%(46/72) & 90.0 (9/10) \\
Other & 80\%(20/25) & No.5 & 50.60\%(42/83) & 80.0 (8/10) \\
\bottomrule
\end{tabular}
}
\label{table_2}
\end{table}

In the single-object tests, each object was grasped five times from different angles to evaluate the GSR.
As shown in Table \ref{table_2}, the results indicate a high success rate across all objects, with an overall GSR exceeding 90\%. 
Certain objects (e.g., Sphere) even achieved a 100\% success rate. This demonstrates that our method is highly reliable in single-object scenes.

For the multi-object scenes, five different camera angles (90°, 75°, 60°, 45°, and 30° relative to the desktop) were used to capture the scene.
Grasping experiments were performed across 10 different object placements, with up to 10 grasp attempts per placement.
After each successful grasp, the object was removed from the scene. As observed from the data in Table \ref{table_2}, the GSR decreases as the camera angle decreases, primarily due to increased occlusion between objects. However, the Scene Completion Rate (SCR) remains relatively stable, which indicates that although occlusion impacts the initial detection of objects, all objects can eventually be grasped after multiple attempts.

In multi-object scenes, occlusions from unfamiliar objects posed additional challenges, as they were not always detected with the most suitable grasp pose. 
However, once an object was successfully grasped and the environment changed or the grasp sequence varied, the system could successfully perform grasping for all objects. 
Furthermore, by relaxing the threshold for the grasp center, the system could achieve successful grasps even in more complex multi-object scenes.

These tests clearly show that the introduction of probabilistic P$n
$P layers enabled us to effectively implement end-to-end learning, allowing the system to better understand depth information. 
Our improved method helps the network better understand 3D data by incorporating 3D supervision signals provided by the real poses on 2D keypoint predictions.
The results strongly support our method's capacity for both high precision and reliable performance across a wide range of scenes, further validating its potential for real-world robotic applications.


\section{Conclusion}

In this paper, we introduced \textit{KGN-Pro}, a novel grasping network that effectively combines the advantages of previous KGNs with enhanced 3D optimization through probabilistic P$n$P layers. 
The network's ability to integrate 3D supervision signals through the 2D keypoint predictions enables end-to-end learning, leading to better generalization across a variety of objects and environments. 
Our method effectively addresses the limitations of previous approaches, such as the reliance on 2D supervision and non-differentiable P$n$P algorithms, by providing a robust, scalable solution for 6-DoF grasp estimation. 
Compared to baseline methods, this innovation substantially improves grasp pose prediction performance, showcasing superior robustness and generalization capabilities.

\bibliographystyle{IEEEtran}
\addtolength{\textheight}{-11cm}
\bibliography{refs}

\end{document}